\documentclass[10pt,twocolumn,letterpaper]{article}

\usepackage{iccv}
\usepackage{times}
\usepackage{epsfig}
\usepackage{graphicx}
\usepackage{caption}
\usepackage{amsmath}
\usepackage{amssymb}
\usepackage{tabularx}
\usepackage{booktabs} 
\usepackage{adjustbox} 
\usepackage{float}
\usepackage{authblk}
\makeatletter
\renewcommand\maketitle{\AB@maketitle} 
\renewcommand\AB@affilsepx{\quad\protect\Affilfont} 
\makeatother

\usepackage[pagebackref=true,breaklinks=true,letterpaper=true,colorlinks,bookmarks=false]{hyperref}

\iccvfinalcopy 

\def\iccvPaperID{10515} 

\ificcvfinal\fi

\begin{document}

\title{Harnessing the Conditioning Sensorium for Improved Image Translation}



\author[1,3]{Cooper Nederhood}
\author[2]{Nicholas Kolkin}
\author[1,3]{Deqing Fu}
\author[1,3]{Jason Salavon}

\affil[1]{The University of Chicago}
\affil[2]{Toyota Technological Institute at Chicago}
\affil[3]{Jason Salavon Studio \authorcr
  \{\tt \small cnederhood, deqing, jsalavon\}@uchicago.edu, \{\tt nick.kolkin\}@ttic.edu}

\twocolumn[{%
\renewcommand\twocolumn[1][]{#1}%
\maketitle

\vspace{-0.75cm}
\begin{center}
    \centering
    \captionsetup{type=figure}
    \includegraphics[width=\linewidth]
    {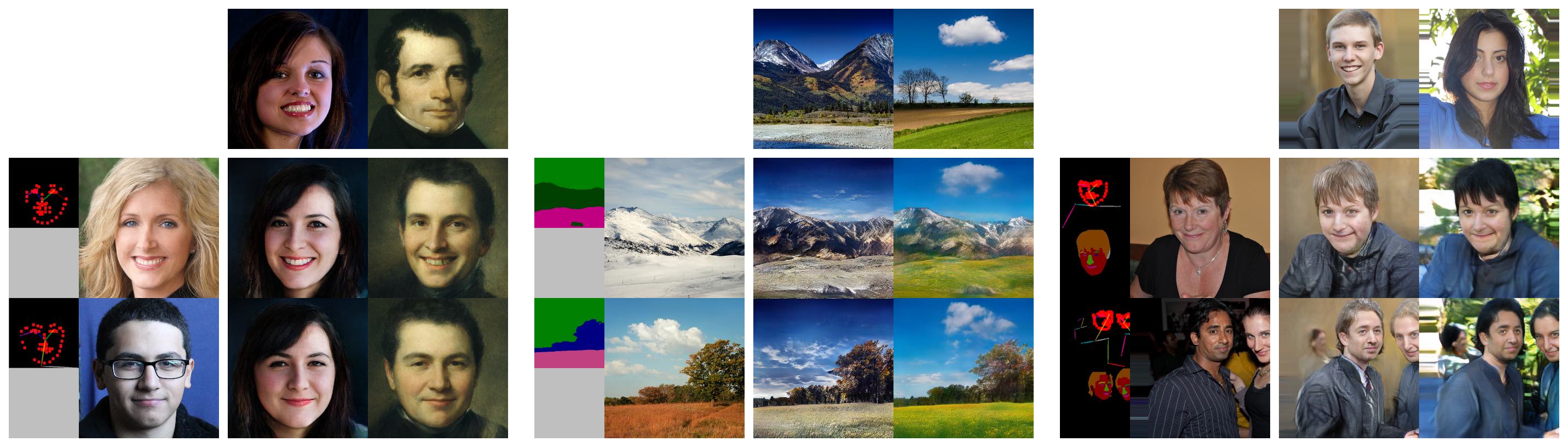}
    \vspace{-0.75cm}
    \captionof{figure}{Image translation results of our framework across a variety of domains and conditioning modalities}
\end{center}%
}]

\maketitle
\ificcvfinal\thispagestyle{empty}\fi

\begin{abstract}
Multi-modal domain translation typically refers to synthesizing a novel image that inherits certain localized attributes from a `content' image (e.g. layout, semantics, or geometry), and inherits everything else (e.g. texture, lighting, sometimes even semantics) from a `style' image. The dominant approach to this task is attempting to learn disentangled `content' and `style' representations from scratch. However, this is not only challenging, but ill-posed, as what users wish to preserve during translation varies depending on their goals. Motivated by this inherent ambiguity, we define `content' based on conditioning information extracted by off-the-shelf pre-trained models. We then train our style extractor and image decoder with an easy to optimize set of reconstruction objectives. The wide variety of high-quality pre-trained models available and simple training procedure makes our approach straightforward to apply across numerous domains and definitions of `content'. Additionally it offers intuitive control over which aspects of 'content' are preserved across domains. We  evaluate our method on traditional, well-aligned, datasets such as CelebA-HQ, and propose two novel datasets for evaluation on more complex scenes: ClassicTV and FFHQ-Wild. Our approach, Sensorium, enables higher quality domain translation for more complex scenes.
\end{abstract}


\section{Introduction}
Our focus in this work is multi-modal, reference-guided, image-to-image domain translation. One well-known variant of this task is seasonal change, for example translating a photograph taken in the summer to a ``winter'' style, adding snow to the ground and removing leaves from trees. However, it's not always clear which features should be invariant across domains. For example, when translating from photographic portraits to paintings a user might wish to either keep their pose or to alter it to match the canon of Renaissance painting. An additional ambiguity arises from the lack of a bijection between most domains. Even within a domain there is typically heterogeneity for a given definition of content. These challenges motivate the development of flexible domain translation models which offer control over what should be preserved as ``content" \textbf{and} which mode of the target domain should be used as ``style".

Existing domain translation frameworks produce high quality results on relatively simple meta-domains like human faces, but performance deteriorates both when the complexity of a single domain increases, and when the semantic and morphological gap between translation domains grows. Rather than pose domain translation as \textit{learning from scratch to disentangle and recombine content and style}, we instead pose it as \textbf{reconstruction from conditioning information, aided by a tightly bottle-necked `style' representation}. Empirically this leads to style representations that capture only what the content does not. Content is defined by selecting conditioning modalities the user wishes to remain invariant during translation (of which a huge variety such as depth-estimation, pose-estimation, and semantic-segmentation, have been well-studied and can be inferred automatically using pre-trained models). This formulation is intuitive, and allows practitioners to select a content definition relevant to their goals.

Our framework, based solely on the well-posed task of image reconstruction, is easy to train and provides the end user with explicit and intuitive control over ``content" while improving synthesis quality for complex scenes. To demonstrate the greater robustness of our pipeline on more complex domain translation tasks we introduce two novel datasets: ClassicTV and FFHQ-Wild. In addition to these complex human-centric translation datasets, we show translations from landscape photographs to different seasons and painting styles.


\section{Related Work}\label{sec:related}
Recent progress in image translation has developed as a specialized subfield of image synthesis. To date the most successful family of neural-network based image synthesis models are Generative Adversarial Networks (GANs). The discriminative loss introduced by GANs has pushed the frontier of high-quality image generation forward, and become a pillar of many image synthesis networks. Autoencoders \cite{rumelhart1985learning}, often combined with discriminative loss, have also proved vital for conditional image synthesis due to their ability to learn to extract data-driven representations from images in an unsupervised manner \cite{zhu2017unpaired,huang2018multimodal, karacan2016learning, park2019semantic}.

While image synthesis from scratch is a challenging and scientifically interesting problem, a clear practical application of image synthesis is as a creative tool. \textbf{Conditional} image synthesis is vital in this context. Numerous forms of spatial conditioning have been explored, for example: class label \cite{brock2018large, mescheder2018training, mirza2014conditional, miyato2018cgans, odena2017conditional}, dense semantic segmentation \cite{karacan2016learning, karacan2018manipulating, zhu2017unpaired, zhu2017toward, Wang2019ExampleGuidedSI}, bounding boxes \cite{zhao2019image}, and pose \cite{esser2018variational,chan2019everybody,Wang2019ExampleGuidedSI, qian2019make}. Not only do conditioned models offer valuable artistic control, they also make complex image synthesis problems tractable. While remarkable progress has been made in unconditional image synthesis, the quality of results in most domains lags far behind that on aligned human faces \cite{Karras_2020_CVPR}. Even high quality models trained using vast amounts of data and computational power\cite{brock2018large,Razavi2019GeneratingDH} are limited to datasets primarily consisting of single, large, central objects such as ImageNet\cite{Deng2009imagenet}, and suffer from a sharp drop in quality relative to models trained only on faces. However, by using ground truth segmentations from the COCO-Stuff \cite{Caesar2018COCOStuffTA} and ADE20k datasets \cite{Zhou_2017_CVPR}, Park et al. \cite{park2019semantic} have recently proposed a method, SPADE, capable of synthesizing outputs of similar quality to \cite{brock2018large,Razavi2019GeneratingDH}, but with significantly greater semantic and spatial complexity. Recent work by Zhang et al. \cite{zhang2020cross} has improved over SPADE, boosting synthesis quality based on ADE20k segmentations even further.

Domain translation has thus far been a specialized form of conditional image synthesis. Prior work has approached this challenge by explicitly learning to disentangle domain invariant and domain specific representations, then mixing them to synthesize a novel image \cite{liu2017unsupervised, park2020contrastive,isola2017image}, learning a bijection between domains by leveraging a cycle consistency loss \cite{zhu2017unpaired}, or by combining both approaches \cite{zhu2017toward, huang2018multimodal}. We propose side-stepping the difficulty of learning domain invariant representations from scratch by taking advantage of automatically generated conditioning information.

Other works have explored conditional image synthesis based on dense \cite{karacan2016learning, karacan2018manipulating, zhu2017unpaired, zhu2017toward, Wang2019ExampleGuidedSI, albahar2019guided} and sparse \cite{zhao2019image,chan2019everybody,Wang2019ExampleGuidedSI, albahar2019guided} spatial inputs, along with simple and domain-specific forms of automatic conditioning such as edges and face parts\cite{zhang2020cross,Wang2019ExampleGuidedSI}. We build on this line of work and demonstrate that even as we move beyond faces to more complex and diverse datasets, such as ClassicTV and FFHQ-Wild, pretrained neural networks are capable of automatically generate high quality conditioning across many modalities. We leverage state-of-the-art models for depth-estimation\cite{lasinger2019towards}, pose-estimation\cite{alp2018densepose}, and semantic-segmentation \cite{he2017mask} to facilitate controllable and high-quality image synthesis.

Not learning a content representation from scratch greatly simplifies the learning process. By using a reconstruction objective we frame style extraction as simply learning to embed all information not contained in the conditioning. To force this information to be spatially invariant we take inspiration from \cite{Karras_2020_CVPR, park2020contrastive} and encode `style' as a vector without spatial dimension extracted from a ``style image''. This vector is injected into our generator as conditional de-normalization \cite{Karras_2020_CVPR}. In fact, we train our domain translation framework solely based on the well-posed task of  image reconstruction. This is in contrast to the more complicated training procedures of recent work, which rely on domain-translation specific regularization terms and losses applied to attempted domain translations (which typically lack a unique solution).
Thus far, domain translation models have primarily focused on translation tasks driven by textural changes such as summer-to-winter or day-to-night \cite{park2020contrastive, isola2017image, huang2018multimodal, chang2020domain, zhu2017unpaired, zhu2017toward, liu2017unsupervised}; or translation within spatially and semantically narrow meta-domains such as aligned human faces/portraits \cite{wu2019disentangling, choi2020stargan, chang2020domain, richardson2020encoding, liu2017unsupervised}, or isolated animals on natural backgrounds\cite{saito2020coco, park2020contrastive,choi2020stargan, huang2018multimodal, chang2020domain, liu2017unsupervised}. We propose two new human-focused datasets for evaluating image translation models: FFHQ-Wild, a dataset of human-centered crops derived from the original in-the-wild FFHQ images; along with ClassicTV, a new dataset composed of central crops from randomly sampled frames of the public domain television shows ``Bonanza'' and ``The Lucy Show'', filtered for frames where human figures are visible. Of the image synthesis models operating in a most similar regime to our own \cite{esser2018variational, lee2018diverse,  huang2018multimodal, Wang2019ExampleGuidedSI, albahar2019guided, choi2020stargan, chang2020domain, zhang2020cross} we compare against the most recent \cite{choi2020stargan, chang2020domain, zhang2020cross} and demonstrate that these models cannot perform well on more spatially complex datasets such as FFHQ-Wild or ClassicTV. We also show that unmodified SPADE \cite{park2019semantic}, which leverages conditioning but is not explicitly designed with domain translation in mind, cannot perform well on these datasets; however, in our ablations we demonstrate that components of our framework can be added to SPADE to improve its performance.

\section{Model}\label{sec:methods}
Our goal is to obtain a model capable of: 
(1) reconstruction, (2) translation within the same domain (guided by an exemplar in the same domain as the input), and (3) translation across domains (guided by an exemplar in the target domain).

 Let $X_a$ and $Y_b$ be two domains, and $x_a$ and $y_b$ be images belonging to $X_a$ and $Y_b$ respectively. Let $c_{a}=\Phi(x_a)$ be the conditioning information extracted by one (or more) pre-trained model(s) from $x_a$. We consider the exact form of $\Phi$ to be a hyperparameter chosen based on the user's goals.
 
There are three components of our model which we train (in contrast to the pre-trained $\Phi$): a content network, a style network, and a generator network. The content network is domain-invariant,  and learns to map the conditioning to a hidden representation $\bar{c_a}=\mathcal{H}(c_a)$ useful for RGB synthesis. All spatial information is contained in $\bar{c_a}$. 

The style network is domain-specific, and maps an RGB image to a global style vector $\bar{s_a} = \mathcal{F}_a(y_a)$ or $\bar{s_b} = \mathcal{F}_b(y_b)$ (depending on the domain of the style). The final output $\hat{x}_{ab} = \mathcal{G}(\bar{c_a},\bar{s_b})$ is produced by the decoder network $\mathcal{G}$ which synthesizes the final image from the spatial conditioning, $\bar{c_a}$, and the global style, $\bar{s_a}$/$\bar{s_b}$. 

If our goal is reconstruction of $x_a$ (an important part of training),  it is produced as $\hat{x}_{aa} = \mathcal{G}(\mathcal{H}(\Phi(x_a)),\mathcal{F}_a(x_a))$. If our goal is style transfer within a domain, it is produced as $\hat{x}_{aa'} = \mathcal{G}(\mathcal{H}(\Phi(x_a)),\mathcal{F}_a(x'_a))$, where $x'_a$ is a second image from domain $X_a$. If our goal is domain translation, it is produced as $\hat{x}_{ab} = \mathcal{G}(\mathcal{H}(\Phi(x_a)),\mathcal{F}_b(y_b))$. Note that this third regime uses a different style encoder $\mathcal{F}_b$, while the first two use the same style encoder $\mathcal{F}_a$.

\subsection{Training objective}
While many domain translation pipelines require complex training procedures, where different sets of losses are applied to different permutations of inputs, our training procedure is very simple. We train our model entirely using content/style representations extracted from the same image, and minimize reconstruction losses in feature space along with an adversarial loss. Because reconstruction is a well-posed and easily optimized task, our model does not suffer from mode collapse, and the adversarial loss encourages the model to hallucinate plausible details not captured by the reconstruction losses, rather then synthesize complex images from scratch.

\textbf{Feature Reconstruction Losses:} We compute our reconstruction loss in the feature spaces of two neural networks. One is pretrained VGG19, and the other is the discriminator being learned (proposed as the GAN-Stability loss \cite{wang2018high}). We use each network to produce activations for the synthesized image and for the ground-truth image. Then, we minimize an L1 loss between hidden layers, weighting the deeper and more semantically meaningful layers more highly.
\begin{equation}
    \small{
    \mathcal L_{ra} = \mathbb  E_{x_a} \sum_{i=0}^{L-1}\left[w_i \left|\left|P^{(i)}\Big(\mathcal{G}(\mathcal{H}(c_a),\mathcal{F}_a(x_a))\Big) - P^{(i)}\Big(x_a\Big)\right|\right|_1 \right]} 
\end{equation}
Where $c_a=\Phi(x_a)$ is the conditioning extracted from $x_a$, and $P$ is either pretrained VGG19 or $D_a$ (the discriminator for domain $X_a$). $P^{(i)}(\cdot)$ indexes the hidden representation of $P(\cdot)$ at layer $i$. Finally, $w_i$ is the weight applied to layer $i$, allowing deeper layers to be weighted more highly. We define the representation loss in domain b, $\mathcal{L}_{rb}$, analogously.

\textbf{Adversarial Loss:} We also use the adversarial loss proposed in Hinge-GAN \cite{lim2017geometric} to encourage synthesizing domain specific details not captured by the reconstruction losses. 
\begin{equation}
\begin{aligned}
    \mathcal L_D^a =& \hspace{0.1cm}\mathbb E_{x_a} [\text{min}(0, -1 - D_a(\mathcal{G}(\mathcal{H}(c_a),\mathcal{F}_a(x_a)))) \\ & \hspace{0.7cm}+ \text{min}(0, -1 + D_a(x_a)) ] \\
    \mathcal L_G^a =& - \mathbb E_{x_a} \left[(D_a(\mathcal{G}(\mathcal{H}(c_a),\mathcal{F}_a(x_a)))\right]
\end{aligned}
\end{equation}
Where $c_a=\Phi(x_a)$ is the conditioning extracted from $x_a$, and $D_a$ is the discriminator for domain $X_a$. 
We emphasize that we only assess loss on the within-domain reconstruction tasks. As in the representation losses,  $\mathcal{L}_{D}^b, \mathcal{L}_{G}^b$, are defined analogously. 

\begin{figure*}[htp] 
\includegraphics[width=\linewidth]{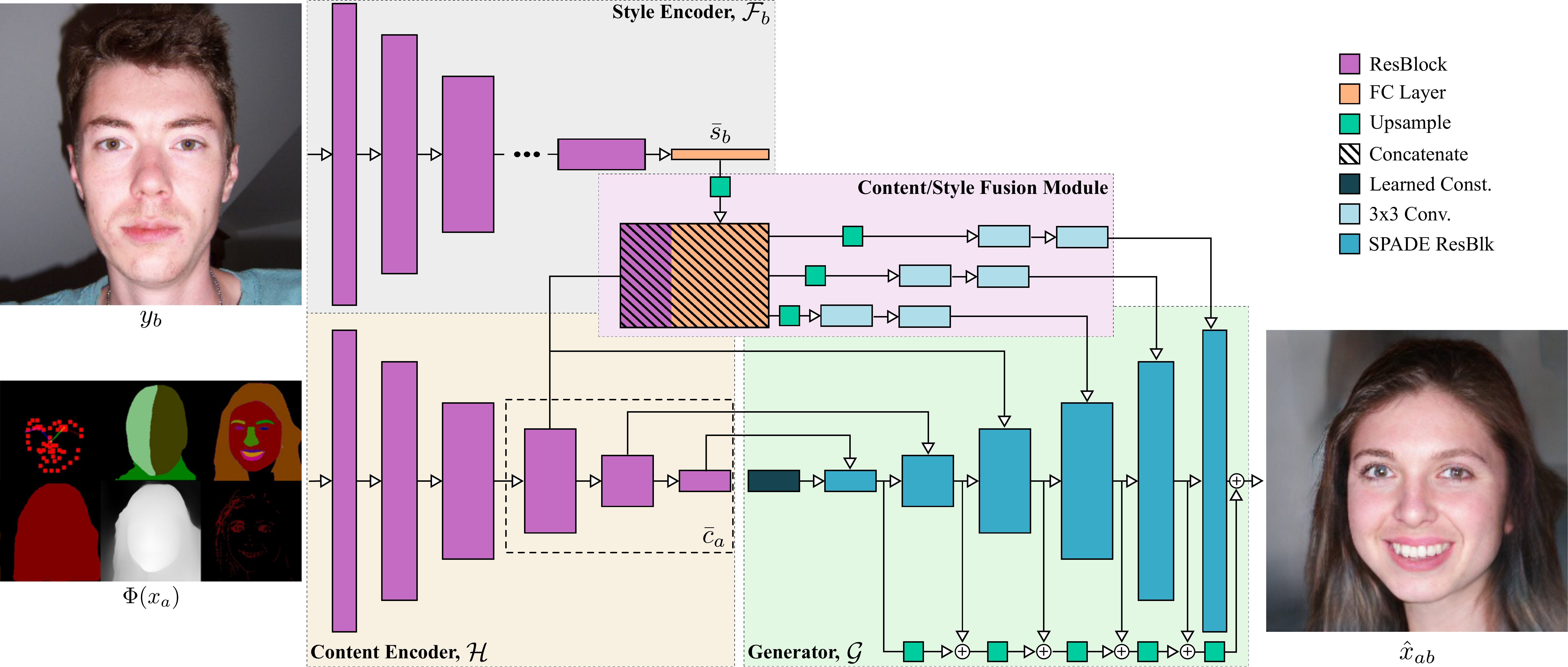}
\caption{Overview of our \textbf{inference} framework. Note that all components of our model are trained purely for reconstruction. Translation from domain $a$ to domain $b$ (as seen here) is never done during training, even for more disparate domains such as FFHQ and Met-Faces.}
\label{fig:generator}
\end{figure*}

\subsection{Architecture}

\textbf{Content Encoder:} The input to the content encoder is a tensor of conditioning information $\Phi(x_a)\in \mathbb{R}^{256\times256\times d}$, where the number of channels $d$ varies based on chosen forms of conditioning. The content encoder, $\mathcal{H}$, is shared across all domains and is composed of a series of residual convolutions and downsampling layers. After each downsampling layer a feature map is emitted at the current resolution, resulting in a feature pyramid capturing the spatial and semantic information that can be extracted from the given content representation. For a more detailed description of the architecture, see Figure \ref{fig:generator}. We refer to the content/conditioning feature map at resolution $r$ as $\bar{c_a}^{(r)}$, and define the set of feature maps to be $\bar{c_a} = \{\bar{c_a}^{(r)} \hspace{0.1cm}|\hspace{0.1cm} r\in\{8,16,32\} \}$. 

\textbf{Style Encoder:} The style encoding networks are the only domain-specific part of our generator (there is also a separate discriminator for each domain). Given image $y_b$ in domain $Y_b$ our style network generates latent code $\bar{s_b} = \mathcal{F}_b(y_b)$, and this is defined analogously for the style encoder of domain $X_a$. Each encoder network is composed of a series of residual blocks and pooling layers which eventually remove the spatial dimensions of the input, after which we apply a final fully-connected layer. See Figure \ref{fig:generator} for more details.

The style encoder learns a global representation which is encouraged to contain all information about the source image not captured by the chosen conditioning sensorium (otherwise reconstruction will be impossible). While this mechanism for reconstructing content and disentangling style is largely successful, it is not perfect. We expect that our framework would benefit from future research improving reconstruction quality and integrating explicit mechanisms for disentangling style and content.

\textbf{Generator Network:} The generator network is a general purpose image synthesis module shared between domains. Agnostic to the original domain of both the source and the style reference image, the generator simply learns to fuse the content signal $\bar{c_a}$ and the style signal $\bar{s_a}$. Past work \cite{karras2019style} show that a learned global style vector can guide image synthesis by channel-wide scale and bias parameters injected via an AdaIn layer \cite{huang2017arbitrary}. In \cite{park2019semantic} the authors introduces the SPADE injection layer, which generalizes the de-normalization of AdaIn to be spatially varying function of a content embedding (itself derived from conditioning such as a semantic label map). 

\begin{figure}[htp]
\includegraphics[width=\columnwidth]{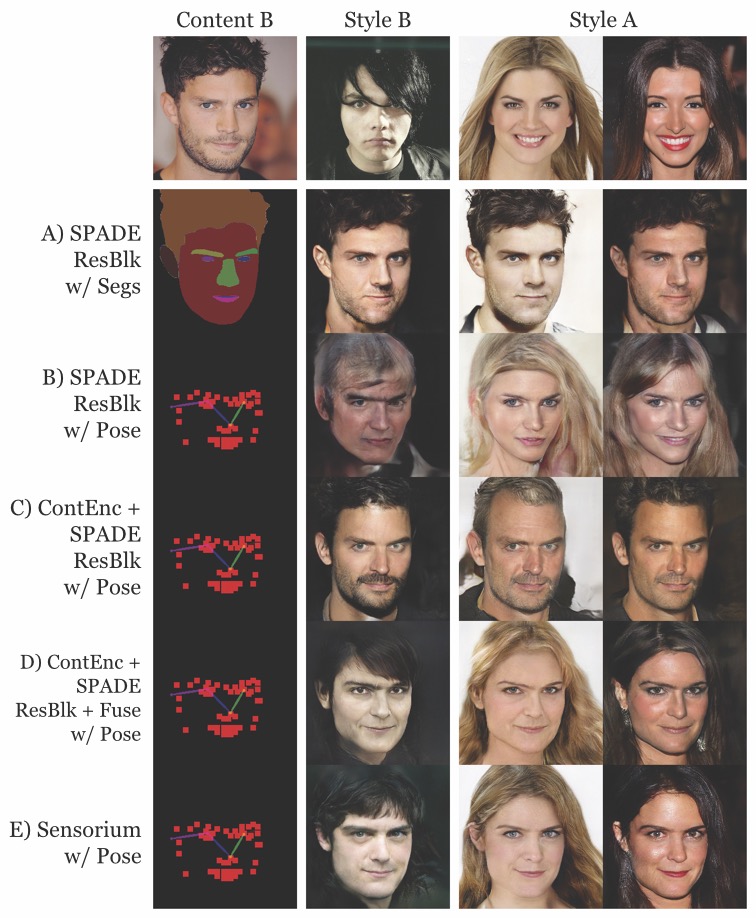}
\caption{Ablation between our method and baseline (SPADE) \cite{park2019semantic}. (A) is SPADE as originally proposed, this synthesizes high quality images when using semantic segmentation as conditioning. (B) is identical to (A) except the conditioning is sparse facial keypoints and head pose, the baseline's performance with sparse conditioning deteriorates significantly. (C) introduces our content encoder, $\mathcal{H}$, before the SPADE residual blocks and image quality approaches that of the original dense conditioning. (D) replaces the VAE based style encoder of \cite{park2019semantic} with our proposed style encoder architecture (shared between domains), $\mathcal{F}$, and Style-Content Fusion module, this dramatically improves how well the output matches the target style. Finally, (E) introduces a separate style encoder and discriminator for each domain ($\mathcal{F}_a,\mathcal{F}_b$, and $\mathcal{D}_a,\mathcal{D}_b$). This emphasizes features characteristics particular to a given domain. For example, note that only (E) recreates the red lips from style reference 3. (E) is our proposed model, Sensorium.}
\label{fig:sensorium_ablation}
\end{figure}

However, as we show in row D of Figure \ref{fig:sensorium_ablation} the style of the output is a dramatically better match to the target exemplar if the scale/bias of SPADE injections are also a function of the style embedding. Before each SPADE injection, we fuse content and style information by spatially replicating our style embedding, then concatenating with the highest resolution feature map in feature pyramid $\bar{c_a}/ \bar{c_b}$. This is followed by bilinear upsampling, then several convolutional layers before predicting the scale/bias parameters for SPADE injection (these layers are learned independently for each SPADE injection). We refer to this as the Content-Style Fusion Module, and it enables the generator to flexibly localize the style information.

\textbf{Discriminator Network:} We adopt the multi-resolution patch-based discriminator from SPADE \cite{park2019semantic}. The discriminator is domain-specific, and we define the discriminators for domains $a, b$ to be $D_a, D_b$ respectively.

\begin{figure}[t]
    \centering
    \includegraphics[width=\linewidth]{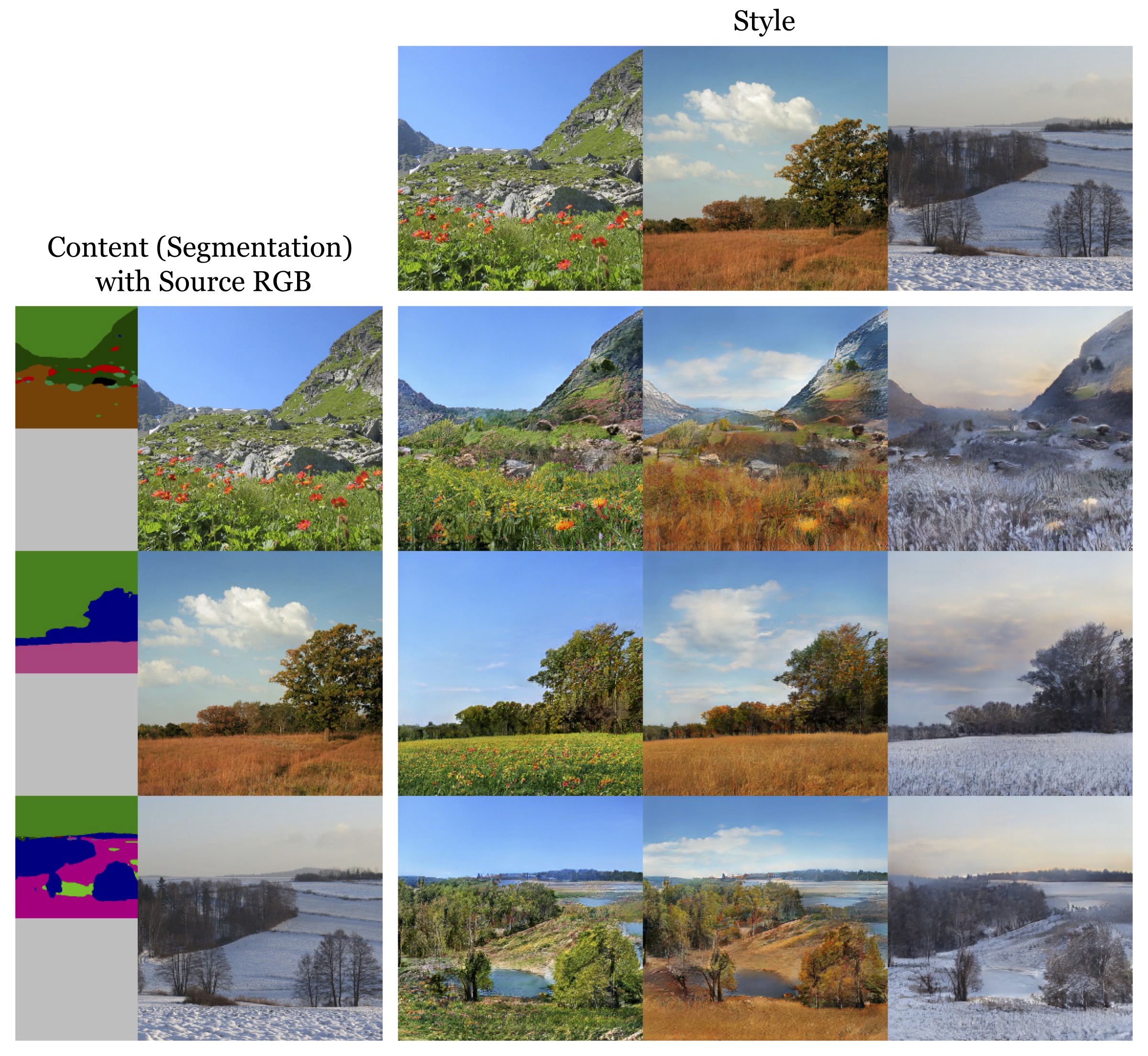}
    \caption{Weight sharing in Sensorium allows for efficient training of multidomain models, like translating between Autumn, Spring, and Winter.}
    \vspace{-0.4cm}
    \label{fig:seasons_translation}
\end{figure}

\section{Evaluation} \label{sec:eval} 
\begin{figure}[tp]
    \centering
    \includegraphics[width=\columnwidth]{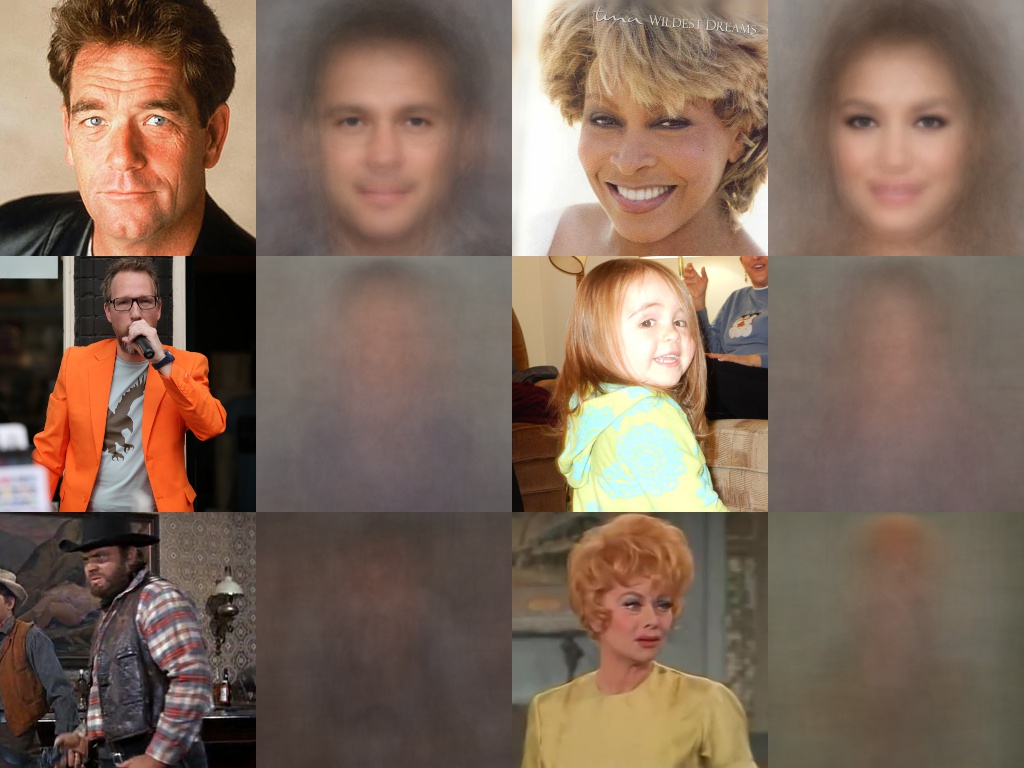}
    \caption{Complexity visualization of evaluation datasets. A random sample from each dataset is shown alongside the mean of all images in that dataset. Row 1 shows CelebA-HQ male and female. Row 2 shows FFHQ-Wild male and female. Finally, row 3 shows ClassicTV Bonanza and LucyShow}
    \label{fig:dataset_viz}
\end{figure}

\begin{figure*}[tp]
    \centering
    \includegraphics[width=\textwidth]{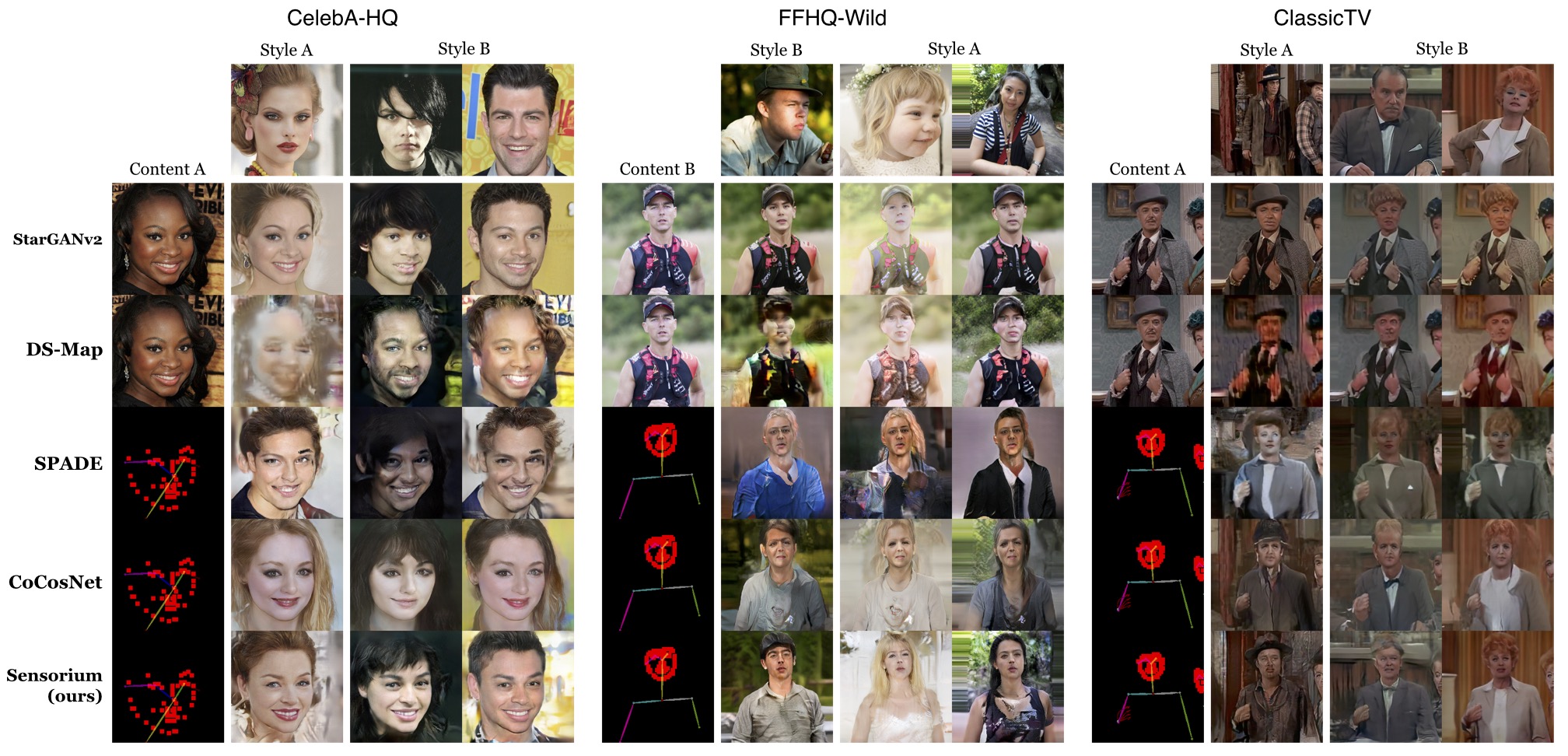}
    \caption{Qualitative comparison of our model and state-of-the-art domain translation models, focusing on the ability to make spatial deformations as dataset complexity increases. From left to right, we picture a content image and then three style reference images. Below the style reference images are synthesized images from StarGANv2 \cite{choi2020stargan}, DS-Map\cite{chang2020domain}, SPADE\cite{park2019semantic}, CoCosNet\cite{zhang2020cross}, and our Sensorium method. We show results for the common task of CelebA-HQ male-female translation where the domains are spatially aligned. We also show results for the spatially varied FFHQ-Wild and ClassicTV datasets. While all methods degrade in quality from CelebA-HQ to FFHQ-Wild and ClassicTV, Sensorium maintains the highest synthesis quality while matching the style well (unlike StarGANv2 and DS-Map, which perform a palette shift).}
    \label{fig:baseline_comp_by_dataset}
\end{figure*}
\begin{figure}[tp]
    \centering
    \includegraphics[width=\linewidth]{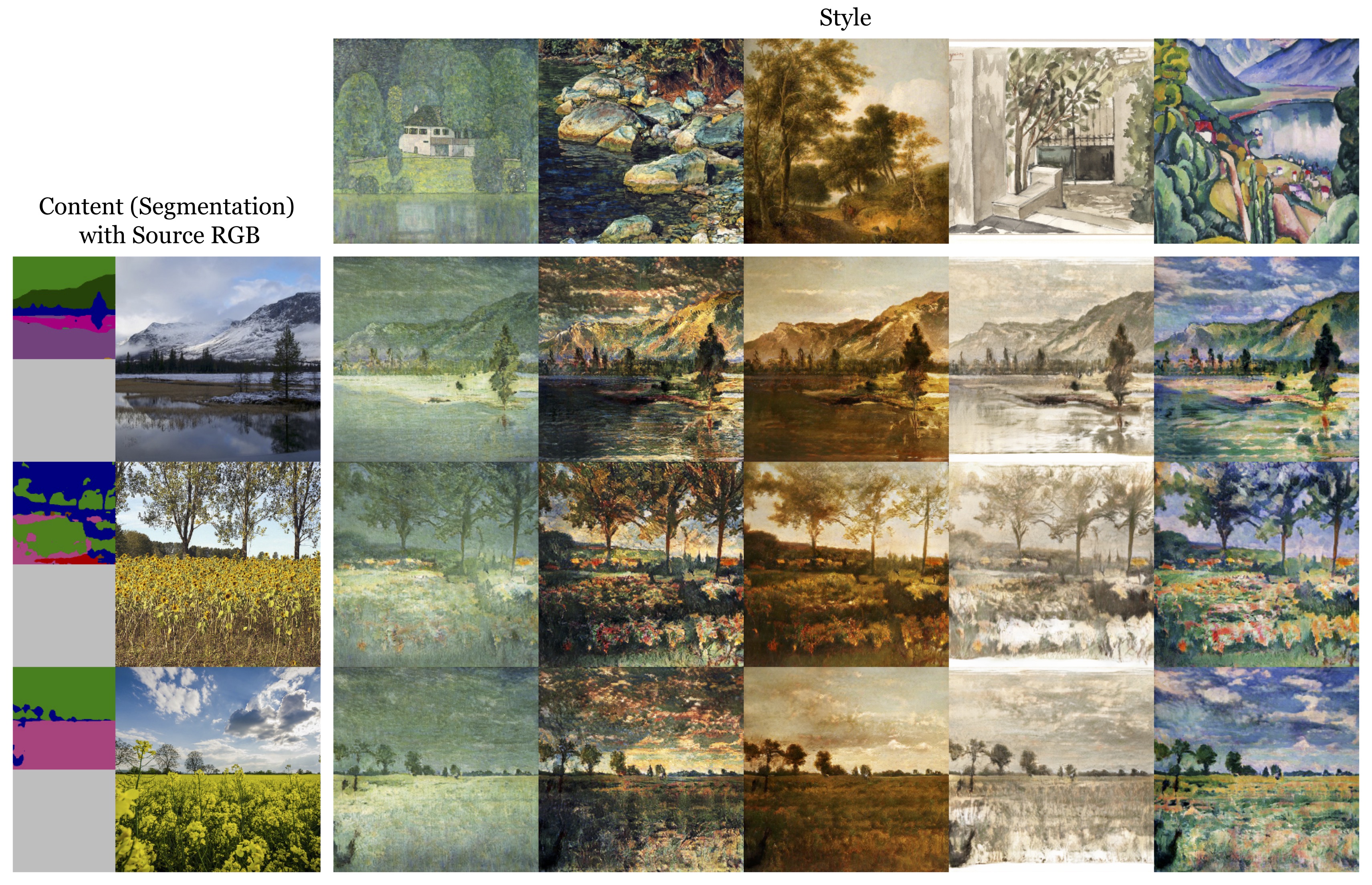}
    \caption{We illustrate performance on the Photos to Paintings task.}
    \vspace{-0.4cm}
    \label{fig:landscape_to_paintings}
\end{figure}

Typical domain translation benchmarks include domain pairs like photograph to painting and summer to winter which highlight textural variations rather than complex morphological differences. We demonstrate performance on these textural-based translation tasks (see Figure \ref{fig:seasons_translation} and Figure \ref{fig:landscape_to_paintings}) and methodically compare on multiple morphologically complex tasks. We benchmark our method on the CelebA-HQ Male/Female datasets and also introduce two novel datasets with much greater spatial variety. We show that performance decays as the spatial complexity increases. In Figure \ref{fig:dataset_viz} we visualize the relative spatial complexity of CelebA-HQ and our two new datasets: FFHQ-Wild Male/Female and ClassicTV Bonanza/LucyShow. 

\subsection{Datasets}
\textbf{CelebA-HQ Male/Female} A dataset of 11,057 high resolution male faces and 18,943 female faces. We hold out 1,000 male faces and 1,000 female faces for validation. 

\textbf{FFHQ-Wild Male/Female} The original FFHQ dataset encompasses 70,000 images of closely cropped human faces. To create a dataset of complex human posses we instead download the original raw in-the-wild images and then use an off the shelf Mask-RCNN model to identify 1-2 human figures from each raw image. We acquire noisey gender labels from the recent FFHQ-Aging dataset \cite{or2020lifespan}. The procedure yields a dataset of 23,462 male crops and 27,305 female crops, from which hold out 238 male crops and 269 female crops for validation. 

\textbf{ClassicTV Bonanza/LucyShow} We also publish a new dataset composed of human figure crops from two classic American television shows: Bonanza and The Lucy Show. Both shows have episodes which have entered the public domain and for each show we randomly sample 40,000 crops from the set of public episodes. To extract a crop from a given frame we use the same process as in FFHQ-Wild, using an off the shelf Mask-RCNN model to identify and crop human figures. For each show, we reserve 10,000 crops sampled from held out episodes. 

\textbf{Flickr Seasons} We qualitatively illustrate performance on the seasonal change problem by downloading season-specific landscape images from Flickr. For Spring, Autumn, and Winter we download approximately 4,000 public images for each domain and we reserve 10\text{\%} for testing purposes. 

\textbf{WikiArt Landscapes} We use 16,000 landscape paintings obtained via the public WikiArt database to learn trainslation from photographic to painterly landscapes. We reserve 1,000 of these images for testing purposes. \cite{tan2017artgan}

\textbf{Derived Conditioning Data} For each dataset we precompute automatically generated conditioning using multiple SOTA models, including: depth \cite{ranftl2019towards}, human pose \cite{8765346} \cite{simon2017hand} \cite{cao2017realtime} \cite{wei2016cpm}, COCO-based segmentations \cite{wu2019detectron2} \cite{he2017mask}, ADE-based semantic segmentations \cite{zhou2018semantic} \cite{zhou2017scene}, facial segmentations \cite{yu2018bisenet}, and human body part segmentations \cite{alp2018densepose}.

\subsection{Baseline Methods}
As discussed, the specifics of domain translation are inherently task dependent. Textural-based translation like Seasonal Change and Photo to Painting are often addressed by style-transfer models whereas more morphological translations like male face to female face and rendering from human pose are often addressed by exemplar-guided image synthesis models. Because we wish to show the robustness of our Sensorium approach to a variety of tasks we compare to both model paradigms, using state-of-the-art approaches for each.

First, as discussed, most domain translation models learn the disentangled style and content representations directly from the RGB images. Originally proposed in MUNIT \cite{huang2018multimodal}, more recent work StarGANv2 \cite{choi2020stargan} and DS-Map \cite{chang2020domain} show improved results. StarGANv2 allows for a style-reference image and sets state-of-the-art performance on the CelebA-HQ benchmark. DS-Map, a style transfer model, maps the content images to a shared space, then further maps the latent representation to domain-specific content spaces, allowing for more expressive content. DS-Map shows performance on a variety of tasks including on the animal faces dog/cat benchmark, which requires morphological changes.

Under our Sensorium approach we allow the choice of derived conditioning to fix the content, allowing style to extract all other image characteristics. Therefore, we also compare Sensorium against models which perform conditional image synthesis with style control via a reference image. SPADE \cite{park2019semantic}, which serves as the starting point for Sensorium, allows for style injection via a reference image. More recently, CoCosNet \cite{zhang2020cross} shows impressive results on exemplar-based image synthesis using a variety of conditioning types including dense semantic segmentations, body pose, and edges.

\subsection{Qualitative Comparison}
Existing methods achieve high-fidelity domain translation when the domains are well aligned. Dataset alignment provides a strong spatial prior for translation. Without this alignment these models lose the ability to make meaningful transformations across domains, and their output devolves to simply palette-matching the style reference. 

In contrast, our method can map complex textural and morphological characteristics from a style reference, while adhering to the layout and semantics provided by the conditioning information. Our approach is more robust to large pose and morphological changes. In Figure \ref{fig:baseline_comp_by_dataset} we show how our approach and prior work behave as the morphological differences between the translation domains grows. Further, while DS-Map and StarGANv2 do not \textit{explicitly} use conditioning data we show that their performance deteriorates  without well-aligned data. This well-aligned data (like for human faces) can only be obtained at scale by employing a pre-trained facial recognition model, essentially the very same models we use to obtain our derived conditioning data. Therefore, our Sensorium approach essentially turns this \textit{implicit} data-processing use of conditioning data into an \textit{explicit} network use of conditioning data. In doing so, our approach becomes more robust to a variety of translation tasks. Please see the supplement for additional results.

\subsection{Quantitative Comparison}
We train each of our three comparison models on our three domain translation datasets: CelebA-HQ M/F; FFHQ-Wild M/F; and ClassicTV. We then train our method using human pose estimated via OpenPose as our initial content representation. We use the Frech\'et inception distance (FID) to quantify performance on the domain transfer task \cite{Heusel2017GANsTB}. To generate the inference population of synthesized images, given a source domain providing content and a (different) target domain providing style reference, we sample 100 images from the validation population for each. We generate all pairwise combinations of content and style reference to yield a total of 10,000 images. Following \cite{choi2020stargan}, we use the training population from the target domain as our ground-truth population  We report the FID figures in Table \ref{tab:fid_table}. 

\begin{table}[htp]
    \scriptsize
    \centering
    \begin{tabularx}{\linewidth}{@{} l c c c c c @{}}
    \toprule 
    &  & & FID $\downarrow$  & & \\
    \cmidrule{2-6} 
        &  DS-Map &  StarGAN-v2 &  SPADE &  CoCosNet &  Ours \\
    \midrule 
    \tiny (Celeb-A) F $\rightarrow$ M & {37.37} & {\color{red} 32.05} & 46.56 & 78.39  & 38.69\\
    \tiny (Celeb-A) M $\rightarrow$ F & {\color{red} 23.81} & { 24.81} & 37.10 & 68.99  & 25.33\\
    \tiny (FFHQ-Wild) F $\rightarrow$ M & { 54.41} & 84.71 & 121.22 & 74.74  & {\color{red} 40.10}\\
    \tiny (FFHQ-Wild) M $\rightarrow$ F & { 61.04} & 121.21 & 152.59 & 98.32  & {\color{red} 45.92}\\
    \tiny Bonanza $\rightarrow$ Lucy & 85.25 & { 83.39} & 137.18 & 60.70  & {\color{red} 57.12}\\
    \tiny Lucy $\rightarrow$ Bonanza & 103.71 & { 83.49} & 130.74 & 67.44  & {\color{red} 47.62}\\
    \midrule 
    Average & { 60.93} & 71.61 & 104.23 & 74.76 & {\color{red} 42.46}\\
    \bottomrule 
    \end{tabularx}
    \caption{Quantitative comparison. We use FID scores (lower is better) to measure the quality of each method on image style translation. Red texts indicate the best method.}
    \label{tab:fid_table}
\end{table}

\vspace{-2ex}
\subsection{Control Via Choice of Content Representation}
\begin{figure*}[htp]
    \centering
    \includegraphics[width=0.95\textwidth]{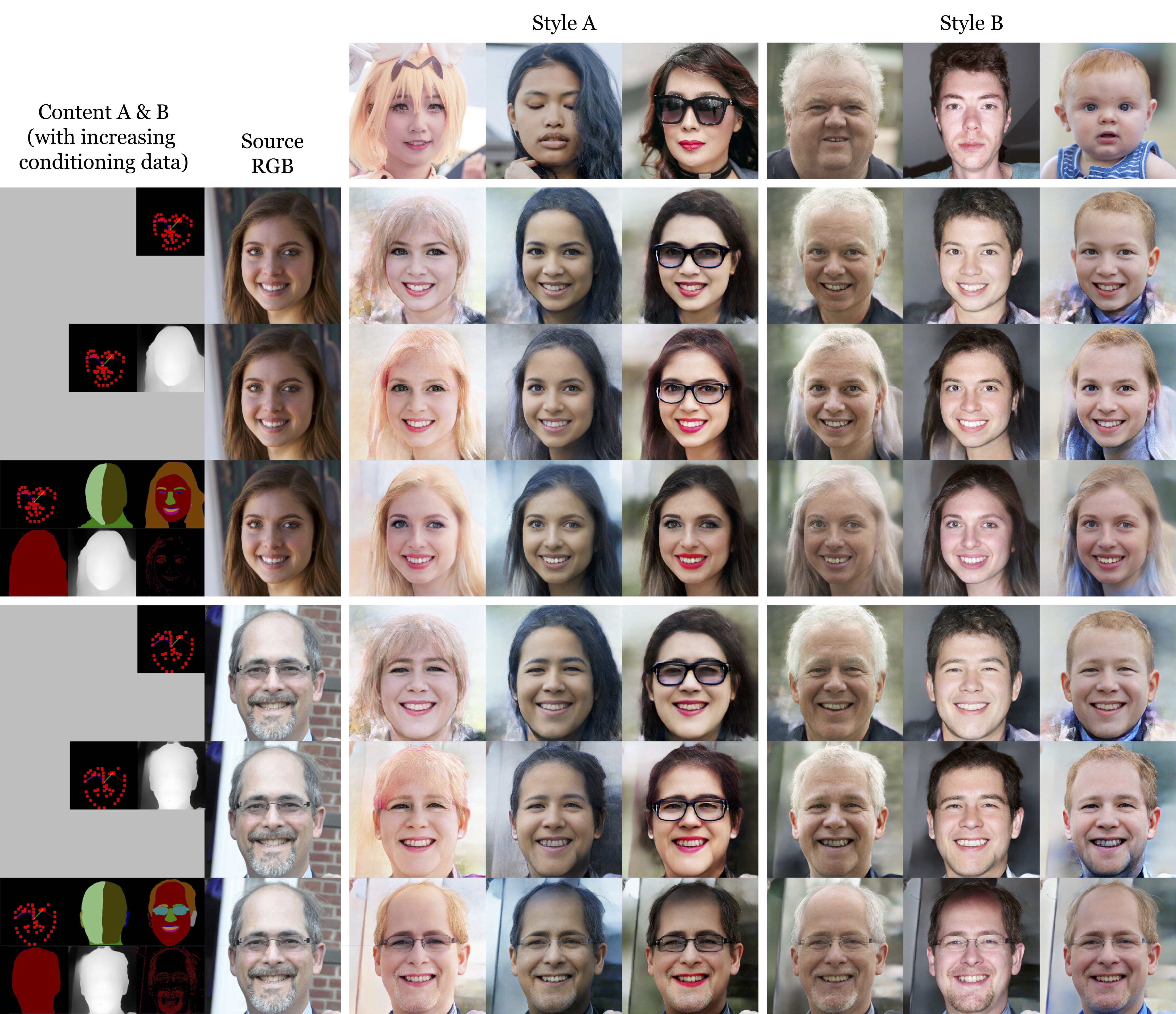}
    \caption{Illustration of the effect on Sensorium of conditioning information with varying levels of detail. Our style encoder learn to extract information needed for reconstruction not contained by the conditioning. As conditioning becomes more detailed the style's effect on the final output decreases. For example, with facial keypoints selected as conditioning Sensorium extracts all remaining geometry including hair and glasses from the style reference image. Adding depth estimation to conditioning then restricts the amount geometry inherited from the style image - hair outline no longer reflects the reference image. Finally, adding facial segmentations fixes features like glasses and edges fix high frequency details like facial hair. }
    \label{fig:dataset_ablation}
\end{figure*}

Our method reliably constrains the synthesized output to match the conditioning information chosen as training inputs. This allows a user to chose the content representation before training (i.e. conditioning information) that suits their goal. For example, a classic style transfer task like transferring the texture of a painting to a photograph calls for holding all geometry constant. In contrast gender transformation tasks typically call for some morphological changes, and conditioning using sparse facial landmarks may be more appropriate. See Figure \ref{fig:dataset_ablation} for detailed examples of how we achieve different degrees of content preservation through the choice of the derived conditioning.

\vspace{-2ex}

\section{Conclusion}
While interesting from a purely scientific perspective, we predict image translation will see rapid growth in usage as a creative tool as user controls are made available. This work explores the use of pretrained models to generate conditioning data at various levels of abstraction, and effectively harness the outputs in an image translation system. 

We show that using automatically generated conditioning allows Sensorium to synthesize compelling domain translations between spatially complex and unaligned domains. In addition, our framework is able to gracefully adjust to different forms of conditioning, filling in aspects of the output the conditioning does not specify using the style of a target exemplar. 

However, tension between identity preservation and domain translation remains a challenge. In particular, our framework can use conditioning
data in conflict with the target domain. Details like ties, beards, and hair will be rendered in the new style whether or not it is appropriate for the target domain. An avenue for future work to address this is developing methods which translate the
conditioning information itself or which integrate global content constraints such as identity. 

Other avenues for future work include extending Sensorium for video translation and a deeper exploration of the relationship between conditioning modalities and corresponding `null-space' controlled by style.

\noindent
\textbf{Acknowledgements} We thank Greg Shakhnarovich for helpful discussion and Jan Brugger for data compilation. This research was supported by the Center for Data and Computing at the University of Chicago and Jason Salavon Studio.

{\small
\bibliographystyle{ieee_fullname}
\bibliography{BIB_oat}
}

\end{document}